\documentclass[11pt]{article}

\usepackage[preprint]{acl}

\usepackage{times}
\usepackage{latexsym}

\usepackage[T1]{fontenc}

\usepackage[utf8]{inputenc}

\usepackage{microtype}

\usepackage{inconsolata}

\usepackage{graphicx}

\usepackage{color}
\usepackage{soul}
\usepackage{amsmath}
\usepackage{amsfonts}
\usepackage{amssymb}
\usepackage{bm}
\usepackage{graphicx}
\usepackage{enumitem}
\usepackage{multirow}
\usepackage{array}
\usepackage{float}
\usepackage{booktabs}
\usepackage{arydshln}
\usepackage{adjustbox}
\usepackage{multicol}
\usepackage{makecell}
\usepackage{xspace}
\usepackage{tcolorbox}
\usepackage[normalem]{ulem}
\usepackage{cleveref}
\usepackage{wrapfig}
\usepackage{xcolor}
\usepackage{afterpage}
\usepackage{caption}
\usepackage{subcaption}
\usepackage{etoolbox}
\usepackage{pifont}

\newtcolorbox{verbatimbox}{
  colback=black!4, 
  colframe=black!4, 
  arc=1pt, 
  boxrule=0.5pt, 
  fontupper=\ttfamily, 
  left=2pt, right=2pt, 
}
\AtBeginEnvironment{verbatimbox}{\fontsize{9}{9}\selectfont}

\crefname{figure}{figure}{figures}
\Crefname{figure}{Figure}{Figures}
\crefname{table}{table}{tables}
\Crefname{table}{Table}{Tables}
\crefname{section}{section}{sections}
\Crefname{section}{Section}{Sections}
\crefname{appendix}{appendix}{appendices}
\Crefname{appendix}{Appendix}{Appendices}

\definecolor{darkblue}{rgb}{0, 0, 0.5}
\hypersetup{colorlinks=true, citecolor=darkblue, linkcolor=darkblue, urlcolor=darkblue}

\definecolor{errred}{HTML}{C9404F}
\definecolor{errpink}{HTML}{E377C2}
\definecolor{erryellow}{HTML}{C9AB40}
\definecolor{errpurple}{HTML}{954CB0}
\definecolor{errgrey}{HTML}{A0A0A0}
\definecolor{errdarkblue}{HTML}{020263}
\definecolor{errbluegray}{HTML}{2F4F4F}
\definecolor{errbrown}{HTML}{8C564B}
\definecolor{errturqoise}{HTML}{1B9E77}
\definecolor{errexaggeration}{HTML}{4B0082}
\definecolor{errfear}{HTML}{008B8B}
\definecolor{errcalling}{HTML}{556B2F}
\definecolor{errbnw}{HTML}{393939}

\newcommand{\errmajor}[1]{{\leavevmode\color{errred} {\ul{\textbf{#1}}}}}
\newcommand{\errminor}[1]{{\leavevmode\color{errpink} {\ul{\textbf{#1}}}}}
\newcommand{\errcontradictory}[1]{{\leavevmode\color{errred} {\ul{\textbf{#1}}}}}
\newcommand{\errmisleading}[1]{{\leavevmode\color{erryellow} {\ul{\textbf{#1}}}}}
\newcommand{\errnotcheck}[1]{{\leavevmode\color{errpurple} {\ul{\textbf{#1}}}}}

\newcommand{\errloaded}[1]{{\leavevmode\color{errbluegray} {\ul{\textbf{#1}}}}}

\newcommand{\errexaggeration}[1]{{\leavevmode\color{errexaggeration} {\ul{\textbf{#1}}}}}
\newcommand{\errfear}[1]{{\leavevmode\color{errfear} {\ul{\textbf{#1}}}}}
\newcommand{\errcalling}[1]{{\leavevmode\color{errcalling} {\ul{\textbf{#1}}}}}

\definecolor{blunavy}{HTML}{0072b2}

\definecolor{orange}{rgb}{0.969, 0.537, 0.008}

\newcommand{\dttask}{\textsc{D2T-Eval}}
\newcommand{\mttask}{\textsc{MT-Eval}}
\newcommand{\proptask}{\textsc{Propaganda}}

\newcommand{\testsplit}{$\mathcal{D}_{\texttt{test}}$}
\newcommand{\devsplit}{$\mathcal{D}_{\texttt{dev}}$}
\newcommand{\iaasplit}{$\mathcal{D}_{\texttt{iaa}}$}

\newcommand{\promptbase}{$\mathcal{P}_{\texttt{base}}$}
\newcommand{\promptnoreason}{$\mathcal{P}_\texttt{noreason}$}
\newcommand{\promptnoguide}{$\mathcal{P}_\texttt{noguide}$}
\newcommand{\promptfewshot}{$\mathcal{P}_\texttt{5shot}$}
\newcommand{\promptcot}{$\mathcal{P}_\texttt{cot}$}

\title{LLMs as Span Annotators: A Comparative Study of LLMs and Humans}

\author{
Zdeněk Kasner$^{1}$ \quad \textbf{Vilém Zouhar}$^{2}$ \quad \textbf{Patrícia Schmidtová}$^{1}$ \quad \\ \textbf{Ivan Kartáč}$^{1}$ \quad
\textbf{Kristýna Onderková}$^{1}$ \quad \textbf{Ondřej Plátek}$^{1}$ \quad \\
\textbf{Dimitra Gkatzia}$^{3}$ \quad \textbf{Saad Mahamood}$^{4}$ \quad \textbf{Ondřej Dušek}$^{1}$ \quad \textbf{Simone Balloccu}$^{5}$ \\
$^{1}$Charles University\quad
$^{2}$ETH Zurich\quad
$^{3}$Edinburgh Napier University\\
$^{4}$trivago N.V.\quad
$^{5}$TU Darmstadt, Germany\\
Contact: \href{mailto:kasner@ufal.mff.cuni.cz}{\texttt{kasner@ufal.mff.cuni.cz}}\\
}

\begin{document}

\maketitle

\begin{abstract}
 Span annotation -- annotating specific text features at the span level -- can be used to evaluate texts where single-score metrics fail to provide actionable feedback. Until recently, span annotation was done by human annotators or fine-tuned models. In this paper, we study whether large language models (LLMs) can serve as an alternative to human annotators. We compare the abilities of LLMs to skilled human annotators on three span annotation tasks: evaluating data-to-text generation, identifying translation errors, and detecting propaganda techniques. We show that overall, LLMs have only moderate inter-annotator agreement (IAA) with human annotators. However, we demonstrate that LLMs make errors at a similar rate as skilled crowdworkers. LLMs also produce annotations at a fraction of the cost per output annotation. We release the dataset of over 40k model and human span annotations for further research.\footnote{Project website: \url{https://llm-span-annotators.github.io}}
\end{abstract}


\section{Introduction}
Fine-grained aspects of texts, such as semantic accuracy or coherence, depend on local lexical choices. To reflect these aspects in quality judgments of texts, techniques are needed that provide the appropriate amount of detail. However, most automatic evaluation metrics for Natural Language Generation (NLG) assign only singular scores for the whole text per each evaluated aspect \cite{gkatzia2015snapshot,sai2022survey,schmidtova2024automatic}. Although numerical values make it easy to rank systems, these metrics are too crude and susceptible to biases or miscalibration of the underlying models \cite{gehrmann2022repairing,liu2023calibrating,wang2023large,gao2024llm}.

\begin{figure*}[t]
  \centering
  \includegraphics[width=1\linewidth]{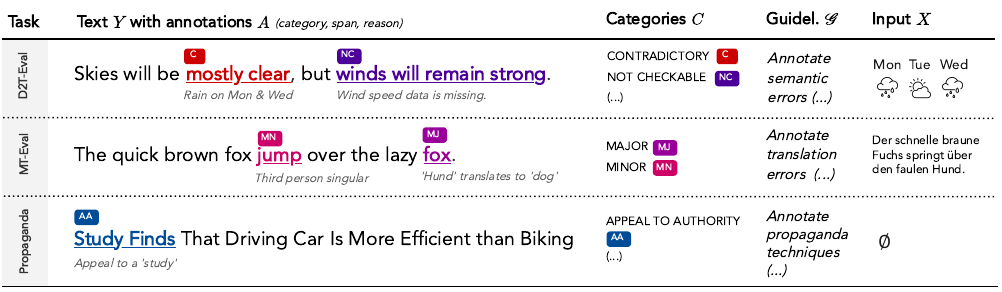}
  \caption{Examples of span annotation tasks that we automate with LLMs. We unify the setup for evaluation tasks (\dttask{}, \mttask{}) and text analysis tasks (\proptask{}).}
  \label{fig:span_eval_tasks}
\end{figure*}

The subject of our study, \emph{span annotation} (\Cref{fig:span_eval_tasks}), offers an alternative approach. Instead of assigning a single score for each evaluated aspect, the goal of span annotation is to localize text spans of interest and classify them according to task-specific guidelines. Span annotations are aligned to specific parts of the evaluated text, which makes them more explainable and actionable than numerical ratings.

Despite its advantages, span annotation has not yet been widely applied in automatic NLG evaluation. The method traditionally required human annotators, making it costly and difficult to scale \cite{dasanmartino2019finegrained,thomson2020gold,popovic2020informative,kocmi2024error}. The \emph{LLM-as-a-judge} paradigm recently emerged as a promising solution to this problem \cite{zheng2023judging,gu2025surveyllmasajudge}, allowing task-specific applications \cite{kocmi2023gembamqm,hasanain2024large}. However, to our knowledge, no study has systematically compared span annotation performance between LLMs and human annotators.

The central focus of our investigation is comparing human annotators and state-of-the-art LLMs on span annotation tasks. We select three span annotation tasks (cf. \Cref{sec:taskdef}): evaluating data-to-text generation \cite{thomson2020gold}, identifying errors in machine translation \cite{kocmi-etal-2024-findings}, and detecting propaganda techniques in human-written texts \cite{dasanmartino2019finegrained}.

Our contributions are as follows:
\begin{enumerate}
  \item We establish that with structured outputs and detailed annotation guidelines, LLMs can serve as robust span annotators, yielding relevant spans for all three annotation tasks we work with (\Cref{sec:automating,sec:res_prompts}).
  \item We show that LLMs have moderate inter-annotator agreement with human annotators overall, but can reach the agreement level among verified crowdworkers who passed a qualification task (\Cref{sec:eval_auto}).
  \item We discover the sources of model errors by manually analyzing a subset of LLM annotation outputs  (\Cref{sec:eval_manual}).
  \item We release a dataset of more than 40k human and model annotations, including annotations collected from crowdworkers and reasoning traces from reasoning LLMs.
\end{enumerate}

\section{Related Work}
\label{sec:rel_work}

\paragraph{LLMs for NLG Evaluation.}

Automatic NLG metrics traditionally assess text quality by measuring similarity to human-written reference texts \cite{sai2022survey,schmidtova2024automatic}. As such, they are unable to quantify more fine-grained aspects \citep{gehrmann2022repairing,10.1162/tacl_a_00437} and do not correlate well with human judgments \cite{novikova-etal-2017-need, reiter-2018-structured}. 
With the emerging LLM-as-a-judge paradigm \citep{gu2025surveyllmasajudge}, LLMs have been applied as evaluators on various tasks, using simple numeric scoring \citep{bavaresco2024llms,liu2023geval, sottana2023evaluation,leiter2023eval4nlp,chiang2023can}, or free-form feedback \citep{li2024generative, kim2024prometheusa, kim2024prometheus,kartac2025openlgauge}. However, as these outputs are not firmly grounded in text, they tend to miss fine-grained aspects and are influenced by LLM biases \cite{stureborg2024large,koo2023benchmarking,wang2023large}.

\paragraph{Span Annotation Protocol.} 
In machine translation (MT), span annotation is a long-standing component of protocols such as MQM or ESA \citep{lommel2014multidimensional,mariana2014multidimensional,popovic2020informative,kocmi2024error}, where human annotators mark erroneous spans in translations. In data-to-text (D2T) generation, span annotation was applied by \citet{thomson2020gold}, who introduced a span-based evaluation protocol for annotation of generated basketball match reports. Span annotation is also used to judge intrinsic text qualities, such as coherence or use of rhetorical devices, in tasks such as propaganda detection \cite{dasanmartino2019finegrained} and text summarization \cite{subbiah2024reading}.  Unlike our work, these works focus on span annotation with human annotators.

\paragraph{Automatic Span Annotation.} Early attempts at automating span annotation with ad-hoc guidelines were based on fine-tuned pre-trained encoder models. That includes evaluation of MT \citep{10.1162/tacl_a_00683}, D2T generation \cite{kasner2021textincontext} or text summarization \cite{goyal2022snac}, as well as propaganda detection \cite{martino2020semeval2020,goffredo2023argumentbased,piskorski2023multilingual}. Automating span annotation with LLMs is more flexible and benefits from increasing LLM capabilities. 
We build on work that applies LLMs as a task-specific evaluation tool \citep{kocmi2023gembamqm,fernandes-etal-2023-devil,hasanain2024large,kasner2024traditional,chang2024booookscore,zouhar2024aiassisted,kartac2025openlgauge,ramponi2025finegrained}. Furthermore, \citet{semin2026strategies} recently investigated various strategies for automatic span annotation with LLMs. Our work is the first that systematically compares the performance of LLMs to human annotators.

\section{Automating Span Annotation with LLMs}

We first formally introduce the span annotation process in \Cref{sec:taskdef}. Next, we discuss how to automate the process with LLMs in \Cref{sec:automating} and how to evaluate the quality of span annotations in \Cref{sec:evaldef}.

\subsection{Span Annotation: Task Definition}
\label{sec:taskdef}
The aim of span annotation is to annotate a \textbf{text sequence $Y = \langle y_1, \ldots, y_{n}\rangle$} given:
\begin{itemize}
  \item the set of \textbf{categories $C = \{c_1, \ldots, c_k\}$},
  \item the annotation \textbf{guidelines $\mathcal{G}$},
  \item the \textbf{source $X$} (such as the translation source; empty if we are annotating only intrinsic text aspects).
\end{itemize}

The output is a set of annotations $A = \{a_1, \ldots, a_m\}$, where each annotation $a_i$ is a tuple $\langle s_i, e_i, c_i, r_i\rangle$:
\begin{itemize}
  \item $s_i, e_i \in \{1,\ldots, n\}, s_i < e_i$ are the start and end indices of the annotated span,
  \item $c_i \in C$ is the assigned annotation category,
  \item $r_i$ is a reason for the annotation (optional).
\end{itemize}

\subsection{Span Annotation with LLMs}
\label{sec:automating}

In our setup, annotations $A$ for the given input $\langle Y, C, \mathcal{G}, X \rangle$ are collected from an LLM:
\begin{align*}
  A = \textsc{LLM}(\text{prompt}(Y, C, \mathcal{G}, X)).
\end{align*}

To obtain the annotations, we follow the setup of \citet{kasner2024traditional}: we request the list of annotations in JSON format, using constrained decoding with a fixed JSON scheme to ensure that the output is syntactically valid. We require each annotation to contain the fields \texttt{reason} (the explanation $r_i$), \texttt{text} (the textual content of the span), and \texttt{type} (the integer index of the error category $c_i$).\footnote{Following \citet{castillo2024structured}, we ensure that the \texttt{reason} field is generated first.}

For reasoning models not supporting structured output, we retrieve the raw answer from the model, strip any parts within the \texttt{<think></think>} tags (if present), and consider the latest valid top-level JSON object as the model's response.

\subsection{Evaluating Span Annotations}
\label{sec:evaldef}

To compare annotations automatically, we need a notion of similarity between two sets of annotations $\mathcal{A}=\{ A_{1}, A_{2}, \ldots, A_{|Y|}\}$ and $\hat{\mathcal{A}} = \{ \hat{A}_{1}, \hat{A}_{2}, \ldots, \hat{A}_{|Y|}\}$ over a set of texts $\mathcal{Y} = \{ Y_{1}, Y_{2}, \ldots, Y_{|Y|}\}$. 
Note that basic inter-annotator agreement metrics such as Cohen's $\kappa$ \citep{cohen1960coefficient} are not applicable in our case, as they require a fixed set of annotation units, while the number and position of spans in the span annotation task may differ \cite{mathet2015unified}.
Therefore, we consider the following similarity metrics:

\paragraph{Pearson correlation $\rho$ over counts.}
This metric compares how many spans were annotated for each example:
\begin{align}
  \operatorname{Pearson}(\mathcal{A}, \hat{\mathcal{A}}) =
  \operatorname{\rho}( \langle |A_{Y}|, |\hat{A}_{Y}|\rangle_{Y \in \mathcal{Y}})
\end{align}
The correlation serves as a sanity check: a low value would suggest that an annotator either skips examples or over-annotates, indicating unclear annotation guidelines.

\paragraph{Precision, Recall, and $\mathbf{F_1}$.} To quantify the degree of alignment between individual annotations, we compute precision, recall, and $F_1$ as defined by \citet{dasanmartino2019finegrained}. These measures are on matching annotations, adjusted to give partial credit to imperfect matches:
\begin{align}
\hspace{-11mm}
\operatorname{Precision}(A_{Y}, \hat{A}_{Y}) = \frac{1}{|A_{Y}|} \sum_{a \in A_Y}  \frac{|a \cap \hat{a}|}{|a|}, \hspace{-8mm}
\end{align}
\begin{align}
  \operatorname{Recall}(A_{Y}, \hat{A}_{Y}) = \frac{1}{|\hat{A}_{Y}|} \sum_{\hat{a} \in \hat{A}_Y}  \frac{|a \cap \hat{a}|}{|\hat{a}|},
\end{align}
where $a \cap \hat{a}$ is the character overlap between two annotation spans and $|a| = e - s + 1$ is the length of the annotation span in characters (see \Cref{sec:taskdef}).
Subsequently, we compute the $F_1$-score as the harmonic mean of precision and recall.

For each of the metrics, we consider \emph{soft} and \emph{hard} variants. The \emph{hard} variant only considers overlaps where the span category matches, while the soft variant disregards the categories. We consider the hard variant to be the default. In addition, we report the difference $F_1\Delta = F_1 (\text{soft}) - F_1 (\text{hard})$.

\paragraph{Gamma $\bm{\gamma}$.}
The $F_1$ score is sensitive to varying span granularities and does not consider near-matches with no overlap or agreement by chance.
To this end, we follow \citet{dasanmartino2019finegrained} and \citet{hasanain2024large} in using the $\gamma$ score \citep{mathet2015unified} as a complementary metric. The metric builds the best possible alignment between the sets of annotations $A_{Y}$ and $\hat{A}_{Y}$ and computes the ``disorder'' of this alignment based on the \emph{positional} and \emph{categorical} dissimilarities of aligned annotations.
The score ranges from $-\infty$ to 1, where 1 is achieved when the annotations are perfectly aligned.
The $\gamma$ score extends Krippendorff's $\alpha$ \citep{krippendorff1980content}, another popular metric, by computing the category-aware span alignments.
We use the implementation of \citet{Titeux2021}.

\paragraph{$S_{\emptyset}$ score.} For an output $y$, one or both annotation sets $A_Y, \hat{A}_Y$ may be empty. This is in fact desirable, e.g., if the goal is to annotate errors in an output that is entirely correct. However, these cases are not properly reflected by the other scores we are using: the F1 score only focuses on counting error spans and is not affected by true negatives, and the $\gamma$ score is undefined if any of the two annotation sets is empty (these examples therefore need to be skipped during the $\gamma$ computation).
To compensate for this, we introduce a score $S_{\emptyset}$that is computed for examples where any of $A_Y$, $\hat{A}_Y$ is empty: 
\begin{align}
  S_{\emptyset} = 1/(1 + |A|),
\end{align}
where:
\begin{align}
|A| = \begin{cases}
|A_Y| & \text{if } |\hat{A}_Y| = 0, \\
|\hat{A}_Y| & \text{otherwise}.
\end{cases}
\end{align}

The score is equal to 1 for the cases where no annotator produced any annotation (i.e., a perfect match) and decreases proportionally to the number of annotations from the annotator that provided a non-zero number of annotations.

\section{Experiments}

\begin{table}[t]
  \centering
  \footnotesize
  \begin{tabular}{l@{\hspace{5pt}}r@{\hspace{5pt}}r@{\hspace{5pt}}r@{\hspace{5pt}}c}
    \toprule
 \textbf{Task}       & \textbf{\# Cat.} & \textbf{\# Texts} & \textbf{Avg. Len}    & \textbf{Novel Data} \\
    \midrule
    \dttask   & 6                & 1,296              & 118/715                         & \ding{52}         \\
    \mttask    & 2                & 2,854             & 26/185                             & \ding{55}          \\
    \proptask & 18               & 100               & 914/4,659                & \ding{55}          \\
    \bottomrule
  \end{tabular}
  \caption{Overview of span annotation tasks used in our experiments. \emph{\# Cat.} denotes the number of categories used in the task (see \Cref{app:categories} for their listings), \emph{\# Texts} the number of texts annotated, \emph{Avg. Len} the average number of words/characters in the output, and \emph{Novel Data} indicates newly collected data.}
  \label{tab:tasks_overview}
\end{table}

\subsection{Tasks}
\label{sec:tasks}
We cover three span annotation tasks of different qualitative aspects. We focus on tasks that do not have extensive training data resources and cannot be readily solved by encoder models (such as, e.g., named entity tagging): evaluating data-to-text generation (\dttask; \Cref{sec:exp:d2t}), identifying errors in machine translation (\mttask; \Cref{sec:exp:mt}), and detecting propaganda techniques (\proptask; \Cref{sec:exp:prop}). See \Cref{tab:tasks_overview} for an overview of our datasets.

\subsubsection{\dttask{}: Evaluation of Data-to-text Generation}
\label{sec:exp:d2t}
In \dttask{}, we use span annotation to evaluate semantic accuracy and stylistic aspects of data-to-text generation outputs \cite{sharma2022innovations,celikyilmaz2021evaluation}. The inputs $X$ are the structured data used to generate the output text $Y$.

We use \dttask{} as a control task to mitigate the effects of \emph{data contamination}: the fact that the performance of the model might be inflated by previous exposure to publicly available benchmarks \cite{balloccu2024leak,dong2024generalization,jiang2024investigating}. Instead of using an existing dataset, we use the \textsc{Quintd} tool \cite{kasner2024traditional} to download structured inputs from multiple public APIs.\footnote{We selected two of the existing domains: \texttt{openweather} for generating weather forecasts and \texttt{gsmarena} for generating phone descriptions. We also add the \texttt{football} domain (using \href{https://rapidapi.com/api-sports/api/api-football}{RapidAPI - API-Football}) for generating soccer game reports.} To obtain output texts for the structured data, we prompt LLMs in a zero-shot setting, asking them to generate a summary of the given data using approximately five sentences. Note that we do not need to deal with the factuality of outputs here, as the sole purpose of the texts is being the input to the annotation process (in fact, having some number of errors is desirable). See Appendix~\ref{app:d2tgen} for more details.

To gather annotations for our dataset, we use crowdworkers from \href{https://prolific.com}{Prolific.com}. We apply best practices for gathering human annotations, including an iterative process to refine annotation guidelines and preselecting the best-performing annotators using a qualification task \cite{tseng2020best,iskender2020best,huang2023incorporating,zhang-etal-2023-needle}.  Our process of collecting annotations proceeded in two stages, following the setup of \citet{zhang-etal-2023-needle}: (1) a \emph{qualification task} for pre-selecting skilled annotators, and (2) the \emph{main task} for collecting the annotations. See Appendix~\ref{app:d2tann} for more details on collecting annotations.

For quality checks, we collect additional internal gold annotation (by the authors) for subsets of the data: \devsplit{} for selecting the best prompt and \iaasplit{} for validating the performance of human annotators (cf. Appendix~\ref{app:d2tann}). Here is a complete overview of our data splits for \dttask{}:
\begin{itemize}[left=0mm,topsep=0mm]
  \item \testsplit{} (1200 outputs) -- for LLM evaluation, annotated with crowdworkers,
  \item \devsplit{} (84 outputs) -- for the study of prompt variants, annotated internally,
  \item \iaasplit{} (12 outputs) -- control for human crowdworkers, annotated internally.
\end{itemize}

\subsubsection{\mttask{}: Identifying Errors in Machine Translation}
\label{sec:exp:mt}
For \mttask{}, we use the dataset of system outputs from the WMT 2024 general shared task \citep{kocmi2024findings}. The system outputs were annotated with the Error Span Annotation (ESA) protocol \citep{kocmi2024error} by professional translators.

The inputs $X$ for \mttask{} are the texts in the source language used to produce the translation $Y$. We follow the WMT 2024 annotation guidelines, focusing on character-level span annotations of \emph{Major} and \emph{Minor} translation errors (see \Cref{tab:cat_mt} for their definitions). Note that unlike for the other tasks, the annotations in \mttask{} cannot overlap and need not be aligned with word boundaries. 

We select the three textual domains present in the WMT 2024 shared task: \texttt{news}, \texttt{literary}, and \texttt{social}; using the data translated from English into other languages: Chinese, Czech, German, Hindi, Icelandic, Japanese, Russian, Spanish, and Ukrainian. 

The original dataset has nearly 50k model outputs, making it too extensive for our evaluation campaign. Therefore, we used a balanced subsample: For each of the nine language pairs, we randomly sample ten input translation segments. We then take all available system outputs for these 90 input segments, making up 2,854 examples in total.

\subsubsection{\proptask{}: Propaganda Technique Detection}
\label{sec:exp:prop}

For the \proptask{} task, we use the dataset of \citet{dasanmartino2019finegrained} containing news collected mostly from on-line propagandistic sources. The token-level annotations in the dataset created by expert annotators cover 18 categories of logical fallacies and persuasion techniques. We use the \texttt{test} split for our experiments. Inputs $X$ are empty for this task, as all annotated categories are intrinsic to the evaluated text $Y$.

\subsection{Collecting LLM annotations}
\label{sec:llmann}

\paragraph{Models} For collecting span annotations with LLMs, we use a mixture of open and proprietary state-of-the-art models:
\begin{itemize}
  \item \textbf{instruction-tuned models}: Llama 3.3 70B \cite{grattafiori2024llama}, GPT-4o \cite{hurst2024gpt}, and Claude 3.7 Sonnet \cite{anthropic2025claude},
  \item \textbf{reasoning models}:\footnote{By \textit{reasoning} models we understand the models that use extra inference time to generate a thinking trace before providing the answer \cite{marjanovic2025deepseek}.} DeepSeek-R1 70B \cite{guo2025deepseek}, o3-mini \cite{openai2025o3mini}, and Gemini 2.0 Flash Thinking \cite{google2025geminithinking}.
\end{itemize}
See \Cref{app:implementation} for details on our experimental setup.

\paragraph{Prompts} We define several prompt variants for our experiments. \promptbase{} is the base prompt that includes the guidelines $\mathcal{G}$ as given to human annotators and asks the model to explain its annotation. By extending \promptbase{}, we implement a few-shot prompt adding 5 examples (\promptfewshot{}) and a chain-of-thought prompt simply asking the model to produce intermediate reasoning (\promptcot{}). We also ablate \promptbase{} by removing extended guidelines (\promptnoguide{}) and not asking for explanations (\promptnoreason{}). The full prompts can be found in \Cref{app:prompts}.

\section{Results}
\label{sec:results}
We first investigate the effect of prompting techniques in \Cref{sec:res_prompts}. Next, we evaluate the LLM annotations using automatic metrics (\Cref{sec:eval_auto}) and manually analyze the errors in the model outputs (\Cref{sec:eval_manual}).

\subsection{Prompting Techniques}
\label{sec:res_prompts}
We perform preliminary experiments on the \dttask{} task \devsplit{} set using open models (Llama 3.3 and DeepSeek-R1) to study the differences between prompting techniques. The results are shown in \Cref{tab:prompting_techniques}.
  \begin{table}[t]
  \centering
  \small
  \setlength{\tabcolsep}{4pt}
  \begin{tabular}{@{}lcccccc@{}}
    \toprule
                      & \multicolumn{3}{c}{\textbf{Llama 3.3}} & \multicolumn{3}{c}{\textbf{DeepSeek-R1}}                                    \\
    \textbf{Prompt}   & $F_1$                                  & $\gamma$                                 & \#a/o & $F_1$ & $\gamma$ & \#a/o \\
    \midrule
    \promptbase{}     & 0.20                                   & 0.13                                     & 2.4   & \textbf{0.25}  & \textbf{0.20}     & 1.0   \\
    \promptcot{}      & 0.09                                   & 0.10                                     & 0.8   & 0.24  & 0.19     & 1.1   \\
    \promptfewshot{}  & \textbf{0.25}                                  & \textbf{0.18}                                     & 2.5   & 0.21  & 0.16     & 1.4   \\
    \promptnoguide{}  & 0.11                                   & 0.08                                     & 3.4   & 0.20  & 0.16     & 1.6   \\
    \promptnoreason{} & 0.22                                   & 0.13                                     & 2.2   & 0.24  & 0.18     & 1.1   \\
    \bottomrule
  \end{tabular}
  \caption{Comparison of prompting techniques on the \devsplit{} (\#a/o is the average number of annotations per output).}
  \label{tab:prompting_techniques}
  \end{table}

  \begin{figure*}[t]
  \centering
  \includegraphics[width=\linewidth]{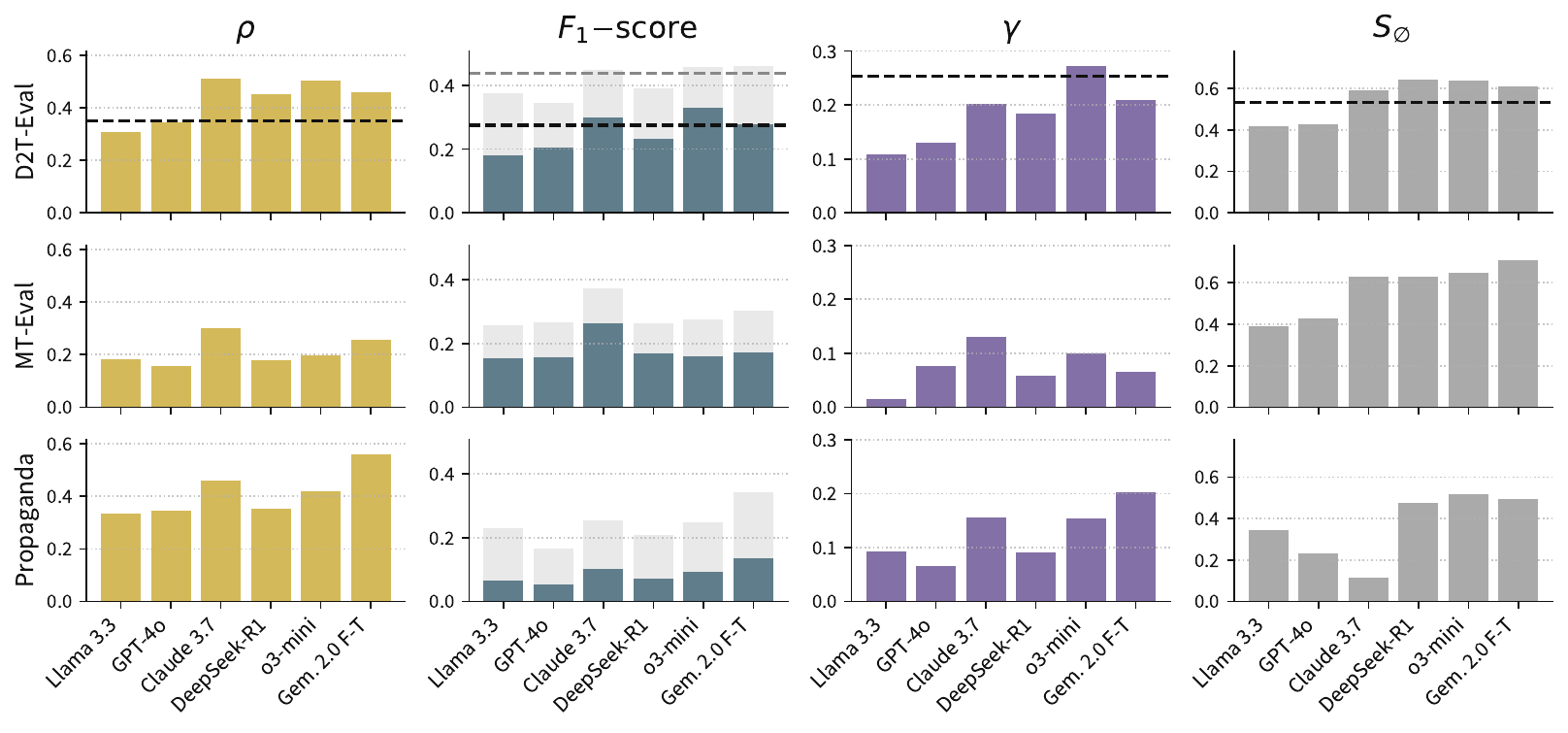}
  \caption{Comparison between LLMs using \promptbase{} and human annotators. Rows represent different tasks (see Section~\ref{sec:tasks}), columns show different annotation similarity metrics (see Section~\ref{sec:evaldef}). For the $F_1$ score, the shadow bar denotes its \emph{soft} variant. The dashed horizontal lines denote agreement between our human annotators for \dttask{} (the agreement is not available for the external datasets). More detailed results are included in \Cref{tab:results-d2t-test-iaa,tab:results-mt-zeroshot-sorted-all-iaa,tab:results-mt-zeroshot-avg-iaa,tab:results-propaganda-zeroshot-iaa}.}
  \label{fig:model_comparison_matrix}
\end{figure*}

Including detailed guidelines seems beneficial: omitting the guidelines (\promptnoguide{}) lowers the performance of both models. In contrast, not letting the model explain the annotation (\promptnoreason{}) does not have a substantial effect. For Llama 3.3, the chain-of-thought (CoT) prompting (\promptcot{}) makes it produce fewer annotations per example than the base variant (0.8 vs.\ 2.4), leading to lower F1 and $\gamma$ scores. Llama 3.3 with \promptcot{} tends to ``overthink'' the annotations, deciding not to annotate cases of errors against which it can find some arguments.

Few-shot prompting (\promptfewshot{}) brings ambivalent results, increasing Llama 3.3 scores but doing the opposite for DeepSeek-R1. This observation is aligned with \citet{guo2025deepseek}, who note that few-shot prompting degrades the performance of DeepSeek-R1. Given these considerations, we decided to use $\mathcal{P}_\texttt{base}$ for further experiments.

\subsection{LLM vs.\ Human Annotations}
\label{sec:eval_auto}

Next, we compare LLM and human annotations using the metrics described in \Cref{sec:evaldef}. The overall results for all tasks are given in \Cref{fig:model_comparison_matrix}. We provide detailed results for individual tasks in the Appendix~\ref{app:results}.

\begin{figure}[t]
  \centering
    \vspace{0pt} 
    \includegraphics[width=\linewidth]{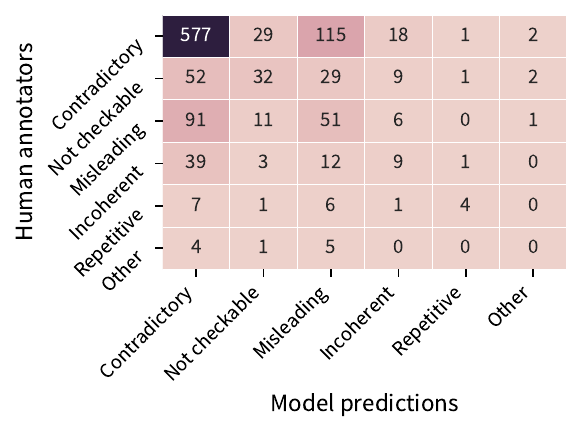}
    \caption{Confusion matrix for \dttask{} (\emph{\underline{C}ontradictory}, \emph{\underline{N}ot checkable}, \emph{\underline{M}isleading}, \emph{\underline{I}ncoherent}, \emph{\underline{R}epetitive}, \emph{\underline{O}ther}), averaged across models (see \Cref{tab:cat_d2t} for category descriptions).}
    \label{fig:confusion_matrix}
\end{figure}

\paragraph{Reasoning models outperform instruction-tuned models.} DeepSeek-R1 generally outperforms the same-sized Llama 3.3.\footnote{The models are comparable as the 70B distilled variant of DeepSeek-R1 is based on Llama 3.3 70B. See \Cref{app:models} for details.} Its superiority is most pronounced on \dttask{} ($F_1$-score of 0.23 vs.\ 0.18, $\gamma$ score of 0.19 vs.\ 0.11). The same observation applies to OpenAI models, where o3-mini outperforms GPT-4o. A notable exception to this trend is the non-reasoning Claude 3.7 Sonnet, which scores mostly on par with o3-mini and excels at \mttask{}.

\begin{table}[t]
  \centering
  \small
  \begin{tabular}{lcc}
    \toprule
    \textbf{Model} & \textbf{Cost (\$/1k out)} & \textbf{Time (s/out)} \\
    \midrule
    crowdworkers          & 500                   & 129.1             \\
    Llama 3.3             & -                     & 21.6              \\
    DeepSeek-R1           & -                     & 227.5             \\
    Claude 3.7 Sonnet     & 10.5                  & 9.0               \\
    o3-mini               & 3.6                   & 21.8              \\
    \bottomrule
  \end{tabular}
  \caption{Estimate of costs and time requirements on \dttask{}: crowdworkers on Prolific, open models (Llama 3.3, DeepSeek-R1), and proprietary models (Claude 3.7 Sonnet, o3-mini).}
  \label{tab:costs}
\end{table}

\paragraph{LLMs reach human IAA on \dttask{}, \proptask{} is harder.} For \dttask{}, we compare model results with an average IAA on a subset of examples annotated by two human annotators. Here, o3-mini, Claude 3.7 and Gemini 2.0 mostly reach or surpass human agreement. For \proptask{}, the upper bound of IAA is the result of \citet{dasanmartino2019finegrained}, who report $\gamma=0.31$ for annotators before consolidation. This score is substantially higher than LLMs (the best LLM score being $\gamma=0.16$ for Claude 3.7 Sonnet). However, this task used expert annotators and has the largest number of categories. The latter property is reflected in the large difference between the soft and hard $F_1$ scores.\footnote{We omit the comparison with human IAA in \mttask{}. While the WMT24 dataset for \mttask{} contains examples annotated with a pair of annotators, these examples take up only a small fraction and exhibit high variance between language pairs.}

\paragraph{Models confuse related categories.} Confusion matrices (see \Cref{fig:confusion_matrix} and \Cref{app:results}) suggest that the models tend mainly to confuse related categories, which may be related to ambiguity or subjective understanding of category definitions. The models also use a less diverse distribution of categories than human annotators.

\paragraph{LLMs are more cost- and time-efficient than human annotators.} An important factor when comparing LLMs and human annotators is efficiency with respect to cost and time per output. For \dttask{}, crowdsourced annotation for 1k outputs costs approximately \$500, while annotating the same amount of outputs with the high-performance o3-mini LLM costs \$3.60 (see \Cref{tab:costs}). In terms of time, the crowdworkers take 129.1 seconds per output on average, which is better than DeepSeek-R1 70B running on our local infrastructure, but an order of magnitude slower than the API-based models.\footnote{Note that we do not ask the crowdworkers to give us a reason $r$ for the annotation, which would arguably make the responses of the crowdworkers slower.} Therefore, LLMs are a more efficient alternative in terms of costs and time.

\begin{figure*}[t]
  \centering
  \includegraphics[width=\linewidth]{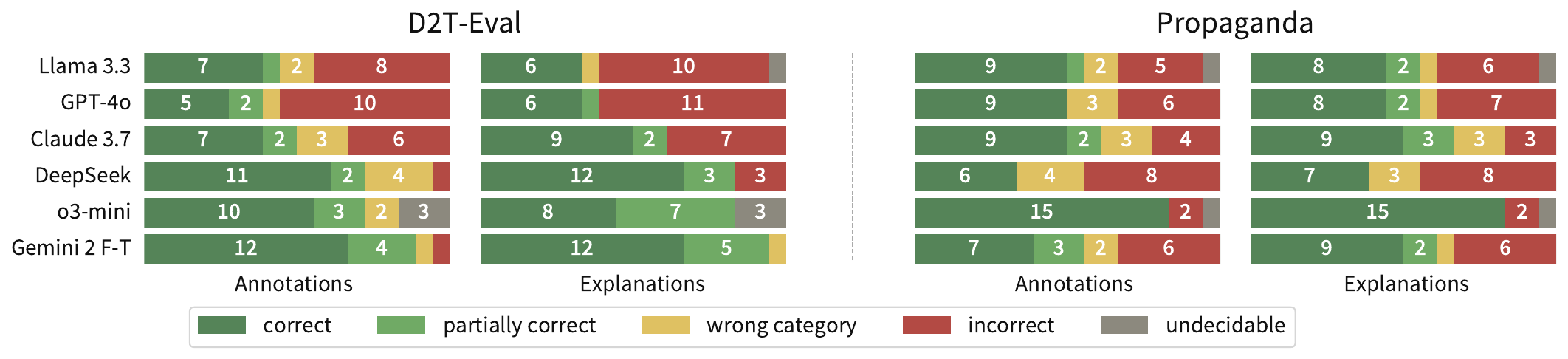}
  \vspace{-1em}
  \caption{Results of our manual analysis. We analyzed 18 annotations and their explanations for each model and task (216 annotations in total). The color bars show annotations that we classified as \emph{Correct}, \emph{Partially correct}, \emph{Wrong category}, \emph{Incorrect}, and \emph{Undecidable}. Detailed results are provided in \Cref{tab:manual-eval-d2t-eval,tab:manual-eval-propaganda}.}
  \label{fig:manual_error_analysis}
\end{figure*}

\subsection{Manual Analysis of LLM Annotations}
\label{sec:eval_manual}
To gain more insights into the qualitative aspects of LLM annotations, we manually analyzed the quality of LLM annotations on 216 samples from \dttask{} and \proptask{}.\footnote{The analysis was split among 7 authors of this paper. While we did not do double annotation due to lack of time, we discussed any unclear cases throughout the process. We do not include \mttask{} in the manual analysis due to our insufficient expertise in most target languages.} For each model, we sampled three annotations per category in \dttask{} and one annotation per category in \proptask{}. Without access to the annotation source, we classified the annotations and their explanations as \emph{Correct}, \emph{Partially correct}, \emph{Wrong category}, \emph{Incorrect}, and \emph{Undecidable}.

We show the results in \Cref{fig:manual_error_analysis} and \Cref{tab:manual-eval-d2t-eval,tab:manual-eval-propaganda}. In total, we marked 49.5\% of LLM-generated annotations and 50.5\% of reasons as correct (with 9.2\% of annotations and 12.5\% of reasons additionally marked as partially correct). Reasoning models perform better, with 56.4\% of their annotations and 58.3\% reasons marked as completely correct. The most accurate annotations on \dttask{} were those made by Gemini 2.0 and DeepSeek-R1. o3-mini performed well on both tasks, although \proptask{} proved challenging for all models. 

\paragraph{What are the sources of model errors?} We find that the models often select wrong error categories despite identifying real issues (e.g., labeling \emph{Contradictory} statements as \emph{Incoherent}). Models also tend to be overly attentive, flagging noise in the data (e.g., markup or off-topic content in \proptask{}) as errors, or marking slight numerical variations (such as rounded values) as misleading. All of these cases could be tackled by more descriptive guidelines or additional examples. However, in some cases, the models also misread or misinterpreted the data (e.g., claiming wind speed measurements do not exist when they do), which hints at deeper issues with understanding the data. Incorrect explanations vary from incomplete explanations (addressing only part of a multi-issue span), irrelevant explanations (e.g., appealing to facts that are ``missing'' from the text) to incorrectly flagging subjective statements (e.g., ``\textit{long-lasting usage}'') as factual errors. Occasionally, the model admits that it marked a correct span as an error, such as in \textit{``The description of the game's duration aligns with the data, providing coherent information''}.\footnote{This typically happened to GPT-4o, even though OpenAI API \href{https://platform.openai.com/docs/guides/structured-outputs?api-mode=chat\#key-ordering}{ensures JSON key ordering} so the explanation \textit{should} have been generated before the annotation (cf. \Cref{sec:automating}).}

\paragraph{How good are human annotations?} Concerningly, the LLM annotations that were marked as correct have only 24\% hard character-level overlap (51\% soft) with human annotations. This fact led us to analyze the quality of human annotations in \dttask{} (the task in which we had the necessary domain expertise). Using the same methodology as we used for the LLM annotations, we annotated a limited sample of 108 human annotations. We marked 45.3\% of the annotations as \emph{Correct}, which is comparable to the LLM annotations (see \Cref{tab:manual-eval-humans} for the results). These findings suggest that the task is hard even for human annotators, and the quality of annotations from crowdworkers varies, even if they are preselected using a qualification task.

\section{Discussion}

Here, we summarize our findings and discuss the implications of our results.

\paragraph{Can LLMs substitute human annotators?} The IAA between LLMs and human annotators is only moderate, suggesting LLMs cannot straightforwardly replace human annotators. However, using LLMs may be a reasonable option in scenarios based on crowdworkers, where the strongest LLMs reach the average IAA between human annotators themselves. In other cases, when deciding whether to employ LLMs as span annotators, one needs to balance desired output quality with other practical aspects. Here, LLMs provide better flexibility, shorter response times, and lower costs. One should also consider the quality of available human annotators, as even qualified crowdworkers (i.e., those who passed a qualification task) make similar amounts of errors as LLMs. It can be also assumed that LLM-based span annotation will benefit from future increases in LLM capabilities, while crowdworkers may increasingly rely on LLM to complete tasks \cite{veselovsky2023artificial}. A promising solution seems to be a hybrid approach in which LLMs pre-annotate the text and humans post-edit the annotations \cite{zouhar2024aiassisted}.

\paragraph{How to deploy LLMs as span annotators?} We recommend providing LLMs with detailed guidelines that describe conventions and how to handle ambiguous cases (cf. \Cref{fig:prompt_d2t_zeroshot}). In contrast, we do not recommend providing specific examples (cf. \Cref{fig:prompt_d2t_fewshot}), as this approach did not bring consistent improvements. Arguably, this is due to the length and complexity of the examples, making them distracting to the model. When using LLMs with custom categories or guidelines, we recommend validating the model's annotations against examples hand-annotated by experts on a sample of the data. In general, reasoning models tend to provide more reliable annotations at the cost of higher response times and token count.

\paragraph{Is the task meaningful despite the low scores?} As pointed out by an anonymous reviewer of this paper, the low annotation accuracy -- as found by our manual analysis -- may indicate a fundamental limitation of the proposed evaluation setup. Span annotation is indeed complex and leaves more room for subjectivity than more straightforward annotation such as simple labels or scores. However, we argue that the detailed actionable feedback gained through span annotation outweighs the increased noise, both in terms of explainability of the outputs and their potential for further processing.

\begin{table}[t]
\centering
\small
\begin{tabular}{@{}lccccc@{}}
\toprule
& \multicolumn{5}{c}{\textbf{Annotations}}\\
Source & C & P & W & I & U \\
\midrule
Human annotators & 49 & 	4 &	17 &	31 &	7  \\
\bottomrule
\end{tabular}
\caption{Manual evaluation results for human annotators on \dttask{}. Categories: C=\emph{Correct}, P=\emph{Partially correct}, W=\emph{Wrong category}, I=\emph{Incorrect}, U=\emph{Undecidable}.}
\label{tab:manual-eval-humans}
\end{table}

\section{Conclusion}

We showed that LLMs can serve as span annotators for three span annotation tasks: evaluating data-to-text generation, identifying errors in machine translation, and detecting propaganda in human-written texts. Our experiments show that LLMs achieve moderate agreement with skilled human annotators. The models perform best in \dttask{}, where they are comparable to verified crowdworkers who passed a qualification task. Reasoning models consistently outperform their instruction-tuned counterparts, delivering more accurate annotations and providing more valid explanations for their decisions. Automating span annotation with LLMs seems to be a promising alternative to fine-grained human evaluation sourced from crowdsourcing platforms, opening the way towards scalable and actionable automatic NLG evaluation methods.

\section*{Limitations} Although we aimed to select a representative sample of models, prompts and tasks, our choice is constrained by our limited time frame and budget. Our estimates of the upper-bound IAA for each task are difficult to establish and depend on many factors, such as the chosen annotation categories, their ambiguity, the annotation guidelines, or the qualification level of human annotators. The estimates are also not readily available for existing datasets and require additional data collection. Due to our insufficient expertise in the target languages, we also do not provide language-specific manual error analysis of results.

As an evaluation method, span annotation is not well-suited for certain NLG evaluation tasks such as annotating omissions or rating the overall text style. In these cases, it is best to combine span annotation with other evaluation methods.

\section*{Author Contributions}
DG and SB first came up with the idea for the project, with SB further coordinating and overseeing the research process. ZK led the experimental design and execution part, including conducting both preliminary and main experiments, organizing the crowdsourcing campaigns, and processing the collected data. Multiple authors (DG, IK, KO, SB, SM, VZ, ZK) participated in the collection of gold data for \dttask{}. Similarly, multiple authors (IK, KO, OD, OP, PS, SB, ZK) were involved in manual evaluation of the model outputs. DG provided financial resources for the Prolific campaigns. SM and SB provided expertise in preparing annotation guidelines and structuring the Prolific campaigns. Data processing and analysis were handled mainly by ZK, VZ, and PS, with VZ providing extra support with the WMT data. The paper was written by ZK, VZ, IK, PS, OD, and SB. 

\section*{Acknowledgments}
This work was funded by the European Union (ERC, NG-NLG, 101039303). It was additionally supported by the National Recovery Plan funded project MPO 60273/24/21300/21000 CEDMO 2.0 NPO and the Charles University Research Centre program No.~24/SSH/009. It used resources of the LINDAT/CLARIAH-CZ Research Infrastructure (Czech Ministry of Education, Youth, and Sports
project No.~LM2023062). 
We thank David M. Howcroft for his early input and contributions to the research methodologies adopted in this study.

\section*{Ethics Statement}
The human evaluation study was approved by the internal ethics committee of our institution. Our human annotators were hired over Prolific and paid the platform-recommended wage of 9 GBP/hour (adjusted to slightly higher rates to account for real annotation times). Annotators were pre-selected on the basis of their primary language (English). All annotators were shown detailed instructions and explanation of the data types, data sources, and the purpose of the research. The domains were selected so that they do not contain sensitive or potentially offensive content. We do not collect demographic data about participants.

\bibliography{bibliography}

\appendix

\crefalias{section}{appendix}

\clearpage

\section{Implementation Details}
\label{app:implementation}

\subsection{Open Models}
\label{app:models}

We run the local models using the \texttt{\href{https://ollama.com/}{ollama}} framework in 4-bit quantization. Specifically, we use \texttt{\href{https://ollama.com/library/llama3.3:70b}{llama3.3:70b}} and \texttt{\href{https://ollama.com/library/deepseek-r1:70b}{deepseek-r1:70b}} (which is based on Llama 3.3 70B) for span annotations. We also use \texttt{\href{https://ollama.com/library/gemma2:2b}{gemma2:2b}} and \texttt{\href{https://ollama.com/library/phi3.5:3.8b}{phi3.5:3.8b}} for generating texts in \dttask{}.




For better reproducibility, we set the seed to 42 and the temperature to 0 for the local models.
We do not use these parameters for proprietary models as these parameters are generally not supported.

We run the models using several GPU variants, including NVIDIA H100 NVL (95G), AMD MI210 (64G), and NVIDIA RTX 3090 (24G).

\subsection{Proprietary Models}
\label{app:models_prop}
We use the following proprietary model versions:
\begin{itemize}
    \item \text{GPT-4o}: \texttt{gpt-4o-2024-11-20}
    \item \text{Claude 3.7 Sonnet}: \texttt{claude-3-7-sonnet-20250219}
    \item \text{o3-mini}: \texttt{o3-mini-2025-01-31}
    \item \text{Gemini 2.0 Flash Thinking}: \texttt{gemini-2.0-flash-thinking-exp-01-21}
\end{itemize}

\subsection{Web Interface}
\label{app:factgenie}
We implement our span annotation process using  \texttt{factgenie} \cite{kasner2024factgenie}: a tool that supports both collecting span annotations from humans via a web interface and from LLMs via API calls.

\Cref{fig:ann_iface} shows samples of our annotation interface implemented in \texttt{factgenie} for human annotators, including data visualizations from the \texttt{football} and \texttt{openweather} domains.

\section{Annotating \dttask{}}

\subsection{Generating Outputs}
\label{app:d2tgen}

For generating the outputs for the structured inputs we collected, we use two larger models -- Llama 3.3 70B \cite{grattafiori2024llama} and GPT-4o \cite{hurst2024gpt} -- and two smaller models -- Gemma 2 2B \cite{team2024gemma} and Phi-3.5 3.8B \cite{abdin2024phi}. See more details on the models in \Cref{app:implementation} and prompts in \Cref{app:prompts}.

\subsection{Collecting annotations}
\label{app:d2tann}

\paragraph{Annotation guidelines}
 For the annotation guidelines, we went through an iterative process to establish the annotation guidelines $\mathcal{G}$ and the annotation categories $C$. We started with a preliminary version of the guidelines and annotation categories, drawing inspiration from the guidelines in previous works \cite{kasner2024traditional,thomson2020gold}. We settled on the following annotation categories (see \Cref{tab:cat_d2t} for details): semantic accuracy errors due to information \emph{Contradictory} to the input, \emph{Not Checkable}, or \emph{Misleading}; any \textit{Incoherent} and \textit{Repetitive} content, and any \emph{Other} errors.

\paragraph{Gold annotations} With the annotation guidelines established, we proceeded to collect our own internally annotated gold data: \devsplit{}, which contains 84 examples annotated individually by one of 7 annotators (12 examples per annotator) and \iaasplit{}, which contains 12 examples annotated commonly by all annotators.\footnote{We selected an example for each of the 4 domains and 3 models.} The purpose of \devsplit{} is to create a high-quality development set for the model prompting study, while the purpose of \iaasplit{} is to pre-select skilled crowdworkers and quantify the performance of crowdworkers during data collection. Our average IAA on \iaasplit{} was $F_1=0.444$ and $\gamma=0.399$.

\paragraph{Crowdsourcing annotations} We gather span annotations for \testsplit{} with crowdworkers from \href{https://prolific.com}{Prolific.com}. Our process of collecting annotations proceeded in two stages, following the setup of \citet{zhang-etal-2023-needle}: (1) a \emph{qualification task} for pre-selecting skilled annotators, and (2) the \emph{main task} for collecting the annotations.

\begin{itemize}
  \item \textbf{Qualification task}: For the qualification task, we pre-selected workers whose first language is English, with >99\% approval rate and more than 100 submissions. The workers were presented with a detailed tutorial with annotation guidelines and examples of individual errors. After the tutorial, we tested the worker performance on five manually pre-selected examples from \iaasplit{}. We invited annotators with the $F_1$ score higher than 0.5 w.r.t. our internal annotations for the main task.
  \item \textbf{Main task}:  Of the 230 annotators who participated in the qualification task, 50 annotators (21.7\%) qualified. Of these, 45 annotators accepted (=90\% turnover rate). For annotating the data in \testsplit{}, we presented each annotator with a batch of 32 examples: 25 examples from \testsplit{} and 7 remaining examples from \iaasplit{} (that is, the examples that we did not use for the qualification task). All the 1200 outputs in \testsplit{} were annotated by at least one annotator. Furthermore, 475 outputs (39.6\%) were annotated by an additional annotator.\footnote{We use examples with two annotators to compute the average IAA for \dttask{} in \Cref{sec:eval_auto}. For other experiments, we use only the outputs from the first annotator as reference data.}
\end{itemize}

For the qualification task, we paid all the annotators an average reward of 9.58 GBP / hour regardless of the qualification outcome. For the main task, we pay all the annotators an average reward of 10.70 GBP / hour. 

\section{Annotation Categories}
\label{app:categories}

\Cref{tab:cat_d2t,tab:cat_mt,tab:cat_prop} show an overview of the annotation span categories that we used for our tasks along with their descriptions.

\begin{table}[htbp]
  \centering
  \small
  \begin{tabular}{@{}p{0.28\columnwidth}p{0.66\columnwidth}@{}}
  \toprule
  \textbf{Category Name} & \textbf{Description} \\ 
  \midrule
  \it Contradictory          & The fact contradicts the data. \\
  \it Not checkable          & The fact cannot be verified from the data. \\
  \it Misleading             & The fact is technically true, but leaves out important information or otherwise distorts the context. \\
  \it Incoherent             & The text uses unnatural phrasing or does not fit the discourse. \\
  \it Repetitive             & The fact has been already mentioned earlier in the text. \\
  \it Other                  & The text is problematic for another reason. \\
  \bottomrule
  \end{tabular}
  \caption{Annotation categories for the \dttask{} task.}
  \label{tab:cat_d2t}
  \end{table}

  \begin{table}[htbp]
    \centering
    \small
    \begin{tabular}{@{}p{0.20\columnwidth}p{0.74\columnwidth}@{}}
    \toprule
    \textbf{Category Name} & \textbf{Description} \\ 
    \midrule
    \it Major                  & An error that disrupts the flow and makes the understandability of text difficult or impossible. \\
    \it Minor                  & An error that does not disrupt the flow significantly and what the text is trying to say is still understandable. \\
    \bottomrule
    \end{tabular}
    \caption{Annotation categories for the \mttask{} task.}
    \label{tab:cat_mt}
    \end{table}

  \begin{table*}[htbp]
    \centering
    \def\arraystretch{1.5}
    \small
    \begin{tabular}{@{}>{\raggedright}p{0.15\textwidth}p{0.8\textwidth}@{}}
      \toprule
      \textbf{Category Name}                        & \textbf{Description}                                                                                                                                                                                                                                                                                                 \\
      \midrule
      \it Appeal to Authority                           & Stating that a claim is true simply because a valid authority or expert on the issue said it was true, without any other supporting evidence offered. We consider the special case in which the reference is not an authority or an expert in this technique, altough it is referred to as Testimonial in literature \\
      \it Appeal to fear-prejudice                      & Seeking to build support for an idea by instilling anxiety and/or panic in the population towards an alternative.    In some cases the support is built based on preconceived judgements                                                                                                                             \\
      \it Bandwagon                                     & Attempting to persuade the target audience to join in and take the course of action because "everyone else is taking the same action"                                                                                                                                                                                \\
      \it Black-and-White Fallacy                       & Presenting two alternative options as the only possibilities, when in fact more possibilities exist. As an the extreme case, tell the audience exactly what actions to take, eliminating any other possible choices (Dictatorship)                                                                                   \\
      \it Causal Oversimplification                     & Assuming a single cause or reason when there are actually multiple causes for an issue.    It includes transferring blame to one person or group of people without investigating the complexities of the issue                                                                                                       \\
      \it Doubt                                         & Questioning the credibility of someone or something                                                                                                                                                                                                                                                                  \\
      \it Exaggeration, Minimisation                    & Either representing something in an excessive manner: making things larger, better, worse (e.g., "the best of the best", "quality guaranteed") or making something seem less important or smaller than it really is (e.g., saying that an insult was just a joke)                                                    \\
      \it Flag-Waving                                   & Playing on strong national feeling (or to any group; e.g., race, gender, political preference) to justify or promote an action or idea                                                                                                                                                                               \\
      \it Loaded Language                               & Using specific words and phrases with strong emotional implications (either positive or negative) to influence an audience                                                                                                                                                                                           \\
      \it Name Calling, Labeling                        & Labeling the object of the propaganda campaign as either something the target audience fears, hates, finds undesirable or loves, praises                                                                                                                                                                             \\
      \it Obfuscation, Intentional Vagueness, Confusion & Using words which are deliberately not clear so that the audience may have its own interpretations.    For example when an unclear phrase with multiple definitions is used within the argument and, therefore, it does not support the conclusion                                                                   \\
      \it Red Herring                                   & Introducing irrelevant material to the issue being discussed, so that everyone``s attention is diverted away from the points made                                                                                                                                                                                    \\
      \it Reductio ad hitlerum                          & Persuading an audience to disapprove an action or idea by suggesting that the idea is popular with groups hated in contempt by the target audience. It can refer to any person or concept with a negative connotation                                                                                                \\
      \it Repetition                                    & epeating the same message over and over again so that the audience will eventually accept it                                                                                                                                                                                                                         \\
      \it Slogans                                       & A brief and striking phrase that may include labeling and stereotyping. Slogans tend to act as emotional appeals                                                                                                                                                                                                     \\
      \it Straw Men                                     & When an opponent``s proposition is substituted with a similar one which is then refuted in place of the original proposition                                                                                                                                                                                         \\
      \it Thought-terminating Cliches                   & Words or phrases that discourage critical thought and meaningful discussion about a given topic. They are typically short, generic sentences that offer seemingly simple answers to complex questions or that distract attention away from other lines of thought                                                    \\
      \it Whataboutism                                  & A technique that attempts to discredit an opponent``s position by charging them with hypocrisy without directly disproving their argument                                                                                                                                                                            \\
      \bottomrule
    \end{tabular}
    \caption{Annotation categories for the \proptask{} task. The categories are adopted from \citet{dasanmartino2019finegrained}.}
    \label{tab:cat_prop}
  \end{table*}

\section{Prompts}
\label{app:prompts}

Here, we provide the model prompts:
\begin{itemize}
  \item  \Cref{fig:prompt_d2t_zeroshot,fig:prompt_d2t_min,fig:prompt_d2t_cot,fig:prompt_d2t_fewshot} show the prompts for the \dttask{} that we use for the experiments in \Cref{sec:res_prompts}.
  \item \Cref{fig:prompt_mt_zeroshot} shows the base prompt we used for \mttask{}.
  \item \Cref{fig:prompt_prop_zeroshot} shows the base prompt we used for \proptask{}.
  \item \Cref{fig:prompt_d2t_gen_football} shows the prompt we used for \emph{generating} the outputs for \dttask{}.
\end{itemize}

\begin{figure}[h]
  \centering
  \footnotesize
  \begin{verbatimbox}
  Your task is to identify errors in the text and classify them. \\
    
  Output the errors as a JSON object with a single key "annotations". The value of "annotations" is a list in which each object contains fields "reason", "text", and "annotation\_type". The value of "reason" is the short sentence justifying the annotation. The value of "text" is the literal value of the identified span (we will later identify the span using string matching). The value of "annotation\_type" is an integer index of the error based on the following list:  \\

  \{categories\} \\

  Examples:  \\
  - Contradictory: The lowest temperature does not drop below 4°C, but the text mentions 2°C.
  - Not checkable: The text mentions that "both teams display aggressive play", which cannot be checked from the data. \\
  - Misleading: The tone of the text suggests there are many sensors out of which just a few are listed here. However, according to the data, the device has exactly these four sensors. \\
  - Incoherent: The text states that both teams had "equal chances until the first half ended scoreless." While this is technically true, the expression sounds unnatural for a sport summary. \\
  - Repetitive: The output text unnecessarily re-states information about a smartphone battery that was mentioned earlier. \\
  - Other: Use this as a last resort when you notice something else not covered by the above categories. \\

  Hints: \\
  - Always try to annotate the longest continuous span (i.e., the whole fact instead of a single word). \\
  - Annotate only the spans that you are sure about. If you are not sure about an annotation, skip it. \\
  - Ignore subjective statements: for example "a lightweight smartphone" highly depends on the context: you should not annotate these statements. \\
  - Numerical conventions: For weather forecasts, we accept both precise numbers (e.g. 10.71°C) and the rounded ones (e.g. 11°C) as long as they agree with the data. \\
  - Annotate only according to your own knowledge. If you are considering using an external tool (Google, ChatGPT etc.), just skip that specific fact. \\
  If there is nothing to annotate in the text, "annotations" will be an empty list. \\

  Given the data: \\
  \`{}\`{}\`{}

    \{data\}

  \`{}\`{}\`{} \\
  annotate the errors in the corresponding text generated from the data: \\
  \`{}\`{}\`{}

    \{text\}

    \`{}\`{}\`{}
  \end{verbatimbox}

  \caption{The prompt \promptbase{} for \dttask{}.}
  \label{fig:prompt_d2t_zeroshot}
\end{figure}

\begin{figure}[h]
  \centering
  \footnotesize
  \begin{verbatimbox}
  Your task is to identify errors in the text and classify them. \\
    
  Output the errors as a JSON object with a single key "annotations". The value of "annotations" is a list in which each object contains fields "reason", "text", and "annotation\_type". The value of "reason" is the short sentence justifying the annotation. The value of "text" is the literal value of the identified span (we will later identify the span using string matching). The value of "annotation\_type" is an integer index of the error based on the following list:  \\

  \{categories\} \\

  Given the data: \\
  \`{}\`{}\`{}

    \{data\}

  \`{}\`{}\`{} \\
  annotate the errors in the corresponding text generated from the data: \\
  \`{}\`{}\`{}

    \{text\}

    \`{}\`{}\`{}
  \end{verbatimbox}

  \caption{The prompt \promptnoguide{} for \dttask{}.}
  \label{fig:prompt_d2t_min}
\end{figure}

\begin{figure}[h]
\centering
\footnotesize
\begin{verbatimbox}
  Think about it step-by-step. You should enclose your chain of thoughts between the <think> and </think> tags. Once you are ready, output the JSON object in the required format. \\

  Example: \\
  \`{}\`{}\`{}

  <think> ... chain of thoughts ... </think> { ... JSON object ... }

  \`{}\`{}\`{}
\end{verbatimbox}

\caption{The additional text added for \promptcot.}
\label{fig:prompt_d2t_cot}
\end{figure}

\begin{figure}[h]
  \centering
  \footnotesize
  \begin{verbatimbox}
    In order to help you with the task, we provide you with five examples of inputs, outputs and annotations: \\

    Example \#1:
    
    data: \\
    \`{}\`{}\`{}
  
      \{data\}
  
    \`{}\`{}\`{} \\
   text: \\
    \`{}\`{}\`{}
  
      \{text\}
  
      \`{}\`{}\`{} \\
    output: \\
    \`{}\`{}\`{}
  
      \{annotations\}
  
      \`{}\`{}\`{}

      (...)
  \end{verbatimbox}

\caption{The additional text added for \promptfewshot.}
\label{fig:prompt_d2t_fewshot}
\end{figure}

\begin{figure}[h]
  \centering
  \footnotesize
  \begin{verbatimbox}
  Your task is to identify errors in the translation and classify them. \\
    
  Output the errors as a JSON object with a single key "annotations". The value of "annotations" is a list in which each object contains fields "reason", "text", and "annotation\_type". The value of "reason" is the short sentence justifying the annotation. The value of "text" is the literal value of the identified span (we will later identify the span using string matching). The value of "annotation\_type" is an integer index of the error based on the following list:\\

  \{categories\} \\

  Error spans can include parts of the words, whole words, or multi-word phrases.

  Hint: errors are usually accuracy-related (addition, mistranslation, omission, untranslated text), fluency-related (character encoding, grammar, inconsistency, punctuation, register, spelling), style-related (awkward), terminology (inappropriate for context, inconsistent use).\\

  Make sure that the annotations are not overlapping. If there is nothing to annotate in the text, "annotations" will be an empty list.\\

  Given the source: \\
  \`{}\`{}\`{}

    \{source\}

  \`{}\`{}\`{} \\
  annotate its translation: \\
  \`{}\`{}\`{}

    \{text\}

    \`{}\`{}\`{}
  \end{verbatimbox}

  \caption{The prompt \promptbase{} for \mttask{}.}
  \label{fig:prompt_mt_zeroshot}
\end{figure}

\begin{figure}[h]
  \centering
  \footnotesize
  \begin{verbatimbox}
    Your task is to identify spans of text that employ propaganda techniques. \\

    Output the errors as a JSON object with a single key "annotations". The value of "annotations" is a list in which each object contains fields "reason", "text", and "annotation\_type". The value of "reason" is the short sentence justifying the annotation. The value of "text" is the literal value of the identified span (we will later identify the span using string matching). The value of "annotation\_type" is an integer index of the error based on the following list: \\

  \{categories\} \\
 
  Now annotate the following text: \\
  \`{}\`{}\`{}

    \{text\}

    \`{}\`{}\`{}
  \end{verbatimbox}

  \caption{The prompt \promptbase{} for \proptask{}.}
  \label{fig:prompt_prop_zeroshot}
\end{figure}

\begin{figure}[h]
  \centering
  \footnotesize
  \begin{verbatimbox}
    Given the structured summary of a football game: \\
    \`{}\`{}\`{}

    \{data\}

    \`{}\`{}\`{}

    Generate a match summary using approximately five sentences. The summary should sound natural, reporting on the important moments of the game. Avoid subjective statements, keep the tone of the summary neutral. Do not fabricate any facts that are not explicitly stated in the data.
  \end{verbatimbox}

  \caption{The prompt used for generating outputs in the \texttt{football} domain for \dttask{}. The prompts for the other domains are analogical. For more robust parsing, we initialize the model response with 'Sure, here is the summary: "' .}
  \label{fig:prompt_d2t_gen_football}
\end{figure}

\begin{figure}[t]
  \centering
  \setlength{\fboxsep}{1pt} 
  \setlength{\fboxrule}{0.5pt} 

  \begin{subfigure}[t]{\columnwidth}
    \centering
    \fcolorbox{gray}{white}{\includegraphics[width=\columnwidth]{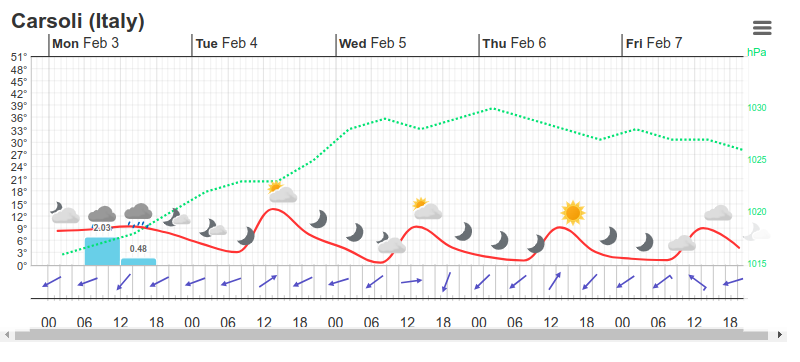}}
    \caption{Data visualization -- \texttt{openweather}}
    \label{fig:ann_iface_b}
  \end{subfigure}
  
  \vspace{0.5em}
  
  \begin{subfigure}[t]{\columnwidth}
    \centering
    \fcolorbox{gray}{white}{\includegraphics[width=\columnwidth]{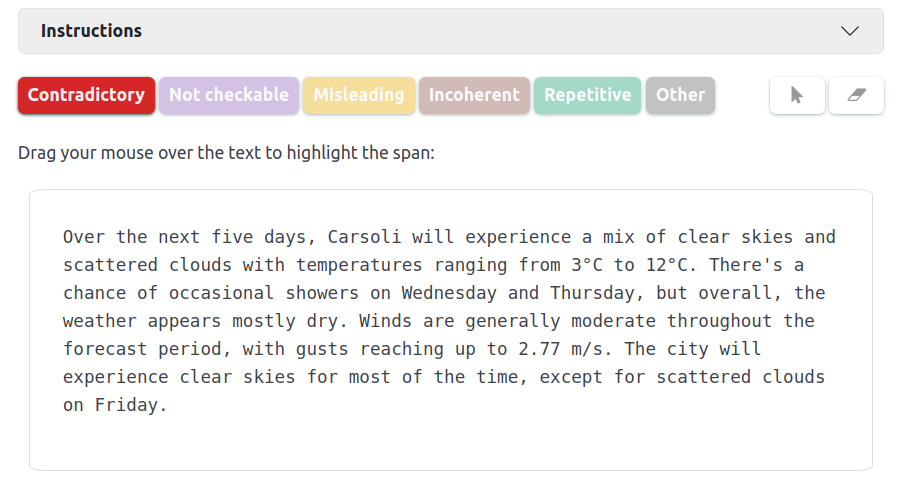}}
    \caption{Interface for highlighting spans}
    \label{fig:ann_iface_c}
  \end{subfigure}
  
  \caption{Samples from the \texttt{factgenie} annotation interface used for collecting span annotations.}
  \label{fig:ann_iface}
\end{figure}

\newcommand{\PreserveBackslash}[1]{\let\temp=\\#1\let\\=\temp}
\newcolumntype{C}[1]{>{\PreserveBackslash\centering}p{#1}}
\newcolumntype{R}[1]{>{\PreserveBackslash\raggedleft}p{#1}}
\newcolumntype{L}[1]{>{\PreserveBackslash\raggedright}p{#1}}

\section{Examples}
\label{app:examples}

In \Cref{tab:example_d2t_weather,tab:example_translation,tab:example_propaganda}, we show examples of the annotated outputs for our tasks. \Cref{fig:ann_iface} shows our annotation interface.

\begin{table*}[htbp]
  \centering
  \footnotesize
  \begin{tabular}{L{2cm}p{13cm}}
    \toprule
   Source (part of the visualization)     & \vspace{1px} \includegraphics[width=1\linewidth]{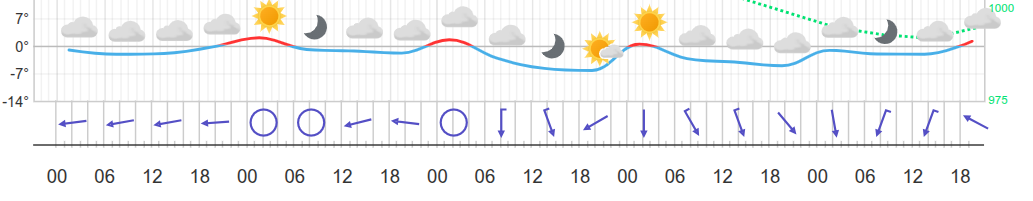}            \\[0em]
Annotated by Llama 3     & Bear Creek will experience mostly overcast skies for the next five days, with \errnotcheck{a chance of occasional showers on Wednesday and Thursday\textsuperscript{not check.}}. Temperatures will remain relatively cool, hovering around freezing point or slightly below. Winds will be light to moderate throughout the period, with \errnotcheck{some gusts expected on Friday\textsuperscript{not check.}}. The overall weather pattern suggests a generally cloudy and potentially wet week. \\[0.8em]

    Annotated by o3-mini     &
   Bear Creek will experience mostly overcast skies for the next five days, with a \errcontradictory{chance of occasional showers on Wednesday and Thursday\textsuperscript{cont.}}. Temperatures will remain relatively cool, hovering around freezing point or slightly below. Winds will be light to moderate throughout the period, \errcontradictory{with some gusts expected on Friday\textsuperscript{cont.}}. The overall weather pattern suggests a generally cloudy and \errcontradictory{potentially wet week\textsuperscript{cont.}}.                                                                            \\[0.8em]

    Annotated by human       &
    Bear Creek will experience mostly overcast skies for the next five days, with a chance \errcontradictory{of occasional showers on Wednesday and Thursday\textsuperscript{cont.}}. Temperatures will remain relatively cool, hovering around freezing point or slightly below. Winds will be light to moderate throughout the period, \errmisleading{with some gusts expected on Friday\textsuperscript{mislead.}}. The overall weather pattern suggests a generally cloudy and \errcontradictory{potentially\textsuperscript{cont.}} \errcontradictory{wet\textsuperscript{cont.}} \errcontradictory{week\textsuperscript{cont.}}.                         \\[0.8em]
    \bottomrule
  \end{tabular}
  \caption{Example for span annotation of \dttask{} in the \texttt{weather} domain with \errcontradictory{contradictory\textsuperscript{cont}}, \errmisleading{misleading\textsuperscript{mislead.}}, and \errnotcheck{not-checkeable\textsuperscript{not check.}} as error span categories. In the annotation interface, the visualization is interactive. The top part (not visible here) shows the place name and date timeline.}
  \label{tab:example_d2t_weather}
\end{table*}

\begin{table*}[htbp]
  \footnotesize \centering
  \begin{tabular}{L{2cm}p{13cm}}
    \toprule

    Source
     & ``It’s your birthday soon, isn’t it?'' Ivory asked, remembering that the princess' birthday was in a few days. Kari nodded, a sad glint in her light blue eyes.                                                                                                                                                                                                                                                                                                                                                                          \\[0.8em]

    Annotated by Llama 3
     & ``¿Es su cumpleaños pronto, \errmajor{no lo es?''\textsuperscript{major}} \errminor{Ivory le preguntó,\textsuperscript{minor}} recordando que el cumpleaños de la princesa era en unos días. Kari nodó, \errmajor{un deslumbramiento triste\textsuperscript{major}} en sus ojos azules claros.                                                                                                                                                                                                                                           \\[0.8em]


    Annotated by o3-mini
     & ``¿Es su cumpleaños pronto, \errminor{no lo es?''\textsuperscript{minor}} Ivory le preguntó, recordando que el cumpleaños de la princesa era en unos días. Kari nodó, un \errmajor{deslumbramiento triste\textsuperscript{major}} en sus ojos azules claros.                                                                                                                                                                                                                                                                             \\[0.8em]

    Annotated by human
     & ``¿Es su cumpleaños pronto, \errminor{no lo es?''\textsuperscript{minor}} Ivory le preguntó, recordando que el cumpleaños de la princesa era en unos días. Kari \errmajor{nodó\textsuperscript{major}}, un deslumbramiento triste en sus ojos azules claros.                                                                                                                                                                                                                                                                             \\

    \bottomrule
  \end{tabular}
  \caption{An example for span annotation of \mttask{} outputs (English$\rightarrow$Spanish) with \errminor{minor} and \errmajor{major} as error span categories.}
  \label{tab:example_translation}
\end{table*}

\begin{table*}[htbp]
  \footnotesize \centering
  \begin{tabular}{L{2cm}p{13cm}}
    \toprule
    Annotated by Llama 3
     &
    \errfear{When the left made Linda Sarsour into its role model, it climbed into bed with\textsuperscript{fear}} \errcalling{the worst of the worst\textsuperscript{labelling}}.
    The father of a missing 4-year-old Georgia boy was training children at \errloaded{a filthy New Mexico compound\textsuperscript{loaded}} to commit school shootings, prosecutors alleged in court documents Wednesday.         \\[0.8em]
    Annotated by o3-mini
     &
    \errloaded{When the left made Linda Sarsour into its role model, it climbed into bed with the worst of the worst.\textsuperscript{loaded}}
    The father of a missing 4-year-old Georgia boy was training children at a  \errloaded{filthy New Mexico compound\textsuperscript{loaded}} to commit school shootings, prosecutors alleged in court documents Wednesday.        \\[0.8em]

    Annotated by human
     & When the left made Linda Sarsour into \errcalling{its role model\textsuperscript{labelling}}, \errloaded{it climbed into bed\textsuperscript{loaded}} \errexaggeration{with the worst of the worst.\textsuperscript{exag.}}
    The father of a missing 4-year-old Georgia boy was training children at \errcalling{a filthy New Mexico compound\textsuperscript{labelling}} to commit school shootings, prosecutors alleged in court documents Wednesday.     \\[0.8em]
    \bottomrule
  \end{tabular}
  \caption{Two examples for span annotation of \proptask{} outputs with \errfear{appeal-to-fear\textsuperscript{fear}}, \errcalling{name-calling-labelling\textsuperscript{labelling}}, \errloaded{loaded-language\textsuperscript{loaded}}, and \errexaggeration{exaggeration\textsuperscript{exag.}} as span categories.}
  \label{tab:example_propaganda}
\end{table*}

\section{Results}
\label{app:results}

\begin{figure}[ht]
  \centering
      \includegraphics[width=0.7\linewidth]{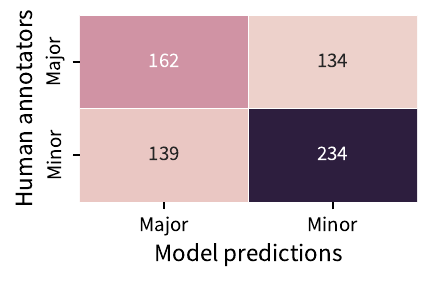}
      \label{fig:confusion_llama_mt}
  \caption{Confusion matrix for \mttask{}, averaged across models (see \Cref{tab:cat_mt} for category descriptions).}
  \label{fig:confusion_matrices_mt}
\end{figure}

\begin{figure*}[ht]
  \centering
    \vspace{0pt} 
    \includegraphics[width=\linewidth]{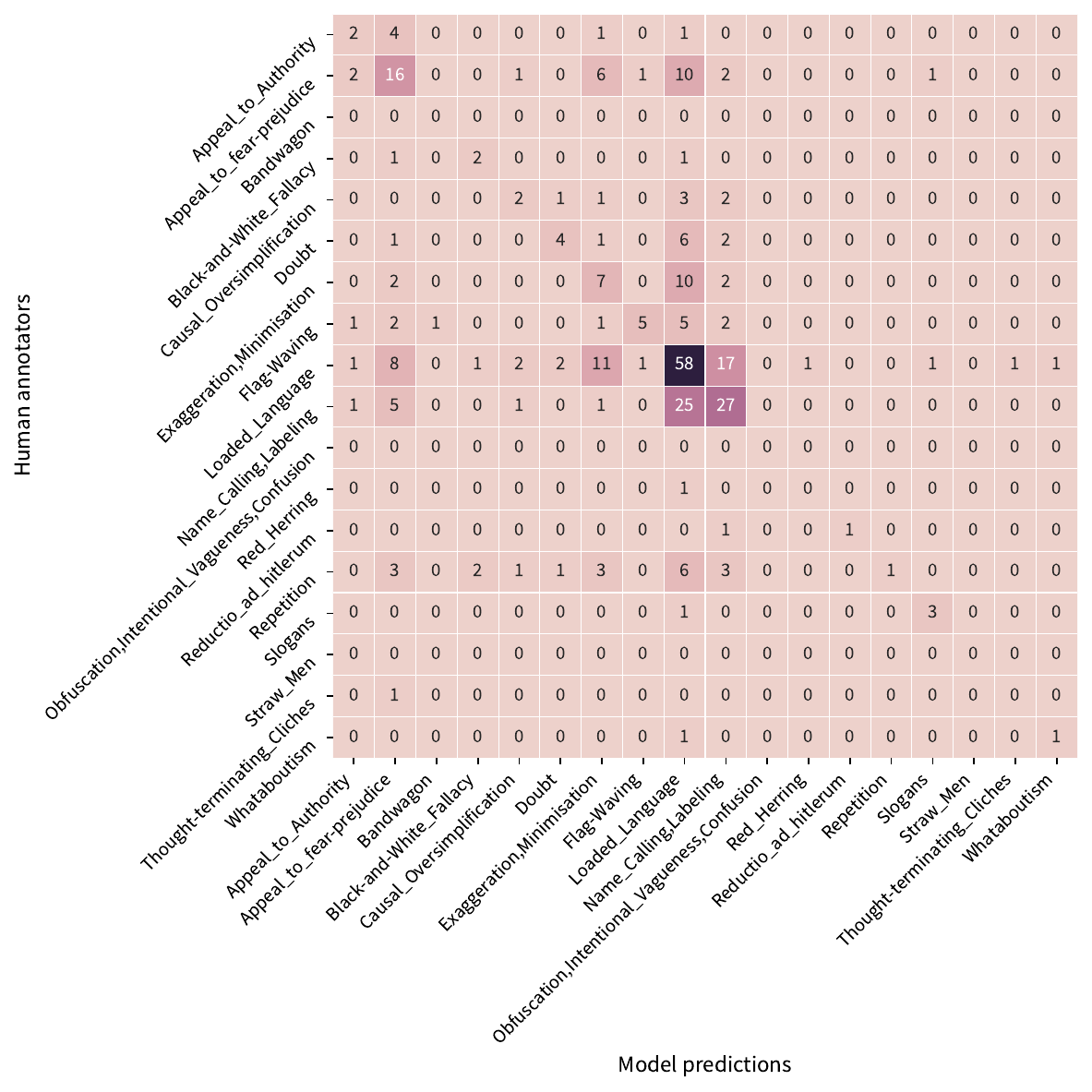}
    \caption{Confusion matrix comparing human annotations (rows) with model predictions (columns) for \proptask{}, averaged across models. (see \Cref{tab:cat_prop} for the description of categories).}
    \label{fig:confusion_matrix_prop}
\end{figure*}

Here, we provide detailed results of our experiments:

\begin{itemize}
    \item \textbf{Main results}: \Cref{tab:results-d2t-test-iaa} (\dttask{}), \Cref{tab:results-mt-zeroshot-sorted-all-iaa,tab:results-mt-zeroshot-avg-iaa} (\mttask{}), \Cref{tab:results-propaganda-zeroshot-iaa} (\proptask{})
    \item \textbf{Extra statistics}: (\dttask{}) \Cref{tab:results-mt-zeroshot-avg-iaa,tab:results-mt-zeroshot-avg-other} (\mttask{}), \Cref{tab:results-propaganda-zeroshot-other}  (\proptask{}).
    \item \textbf{Confusion matrices}: \Cref{fig:confusion_matrices_mt} (\mttask{}) and \Cref{fig:confusion_matrix_prop} (\proptask{}).
    \item \textbf{Manual evaluation}: \Cref{tab:manual-eval-d2t-eval} (\dttask{}), \Cref{tab:manual-eval-propaganda} (\proptask{}), \Cref{tab:manual-eval-humans} (human annotators).
\end{itemize}

\label{sec:res_1}

\begin{table*}[htbp]
\centering

\begin{tabular}{@{}lcccccccccc@{}}
\toprule
\textbf{Model} & \textbf{$\rho$} & \multicolumn{2}{c}{\textbf{Precision}} & \multicolumn{2}{c}{\textbf{Recall}} & \multicolumn{3}{c}{\textbf{F1}} & \textbf{$\gamma$} & \textbf{$S_{\emptyset}$} \\
 &  & Hard & Soft & Hard & Soft & Hard & Soft & $\Delta$ &  &  \\
\midrule
Llama 3.3 & 0.307 & 0.176 & 0.365 & 0.187 & 0.388 & 0.181 & 0.377 & 0.196 & 0.109 & 0.418 \\
GPT-4o & 0.346 & 0.233 & 0.391 & 0.184 & 0.308 & 0.206 & 0.345 & 0.139 & 0.130 & 0.429 \\
Claude 3.7 & \textbf{0.512} & 0.294 & 0.442 & \textbf{0.304} & \textbf{0.457} & 0.299 & 0.449 & 0.150 & 0.203 & 0.592 \\
DeepS. R1 & 0.453 & 0.317 & 0.532 & 0.185 & 0.310 & 0.233 & 0.392 & 0.159 & 0.185 & \textbf{0.645} \\
o3-mini & 0.505 & \textbf{0.392} & \textbf{0.542} & 0.285 & 0.395 & \textbf{0.330} & 0.457 & 0.127 & \textbf{0.273} & 0.637 \\
Gem. 2-FT & 0.458 & 0.293 & 0.488 & 0.263 & 0.438 & 0.277 & \textbf{0.462} & 0.185 & 0.209 & 0.612 \\
\bottomrule
\end{tabular}
\caption{Evaluation of human and LLM annotations using \promptbase{} on \dttask{}. See \Cref{fig:model_comparison_matrix} for visualizaton of this table.}
\label{tab:results-d2t-test-iaa}
\end{table*}

\label{sec:res_5}
\begin{table*}[htbp]
    \centering
    
\begin{tabular}{@{}lcccccccccc@{}}
\toprule
\textbf{Model} & \textbf{$\rho$} & \multicolumn{2}{c}{\textbf{Precision}} & \multicolumn{2}{c}{\textbf{Recall}} & \multicolumn{3}{c}{\textbf{F1}} & \textbf{$\gamma$} & \textbf{$S_{\emptyset}$} \\
 &  & Hard & Soft & Hard & Soft & Hard & Soft & $\Delta$ &  &  \\
\midrule
Llama 3.3 & 0.182 & 0.121 & 0.200 & 0.229 & 0.378 & 0.155 & 0.257 & 0.102 & 0.014 & 0.392 \\
GPT-4o & 0.158 & 0.141 & 0.240 & 0.195 & 0.327 & 0.156 & 0.266 & 0.110 & 0.076 & 0.428 \\
Claude 3.7 & \textbf{0.301} & \textbf{0.226} & \textbf{0.325} & \textbf{0.335} & \textbf{0.469} & \textbf{0.262} & \textbf{0.373} & 0.111 & \textbf{0.131} & 0.628 \\
DeepS. R1 & 0.177 & 0.169 & 0.268 & 0.183 & 0.280 & 0.168 & 0.262 & 0.094 & 0.058 & 0.631 \\
o3-mini & 0.197 & 0.169 & 0.291 & 0.161 & 0.275 & 0.160 & 0.275 & 0.115 & 0.100 & 0.646 \\
Gem. 2-FT & 0.257 & 0.184 & 0.312 & 0.180 & 0.339 & 0.173 & 0.304 & 0.130 & 0.066 & \textbf{0.710} \\
\bottomrule
\end{tabular}
\caption{Evaluation of human and LLM annotations using \promptbase{} on the \mttask{} -- average across languages.}
\label{tab:results-mt-zeroshot-sorted-all-iaa}
\end{table*}

\label{sec:res_mt_lang}

\begin{table*}[htbp]
    \centering
    
\begin{tabular}{@{}lcccccccccc@{}}
\toprule
\textbf{Model} & \textbf{$\rho$} & \multicolumn{2}{c}{\textbf{Precision}} & \multicolumn{2}{c}{\textbf{Recall}} & \multicolumn{3}{c}{\textbf{F1}} & \textbf{$\gamma$} & \textbf{$S_{\emptyset}$} \\
 &  & Hard & Soft & Hard & Soft & Hard & Soft & $\Delta$ &  &  \\
\midrule
en-cs & 0.303 & 0.144 & 0.268 & 0.180 & 0.326 & 0.156 & 0.286 & 0.130 & 0.084 & 0.582 \\
en-es & 0.171 & 0.161 & 0.243 & 0.236 & 0.362 & 0.190 & 0.288 & 0.098 & 0.080 & \textbf{0.631} \\
en-hi & 0.170 & 0.173 & 0.265 & 0.208 & 0.327 & 0.173 & 0.269 & 0.096 & -0.0 & 0.552 \\
en-is & \textbf{0.347} & 0.136 & 0.246 & 0.187 & 0.361 & 0.145 & 0.269 & 0.124 & 0.108 & 0.493 \\
en-ja & 0.127 & 0.193 & 0.302 & 0.249 & 0.363 & \textbf{0.209} & \textbf{0.318} & 0.109 & 0.063 & 0.569 \\
en-ru & 0.225 & 0.178 & 0.256 & \textbf{0.273} & \textbf{0.386} & 0.208 & 0.298 & 0.090 & \textbf{0.162} & 0.588 \\
en-uk & 0.192 & 0.166 & 0.254 & 0.214 & 0.339 & 0.184 & 0.286 & 0.102 & 0.031 & 0.542 \\
en-zh & 0.163 & \textbf{0.196} & \textbf{0.346} & 0.163 & 0.294 & 0.169 & 0.302 & 0.133 & 0.075 & 0.623 \\
\bottomrule
\end{tabular}
\caption{Evaluation of human and LLM annotations using \promptbase{} on the \mttask{} separately for each language (average across models).}
\label{tab:results-mt-zeroshot-avg-iaa}
\end{table*}

\label{sec:res_6}
\begin{table*}[htbp]
    \centering
    
\begin{tabular}{@{}lcccccccccc@{}}
\toprule
\textbf{Model} & \textbf{$\rho$} & \multicolumn{2}{c}{\textbf{Precision}} & \multicolumn{2}{c}{\textbf{Recall}} & \multicolumn{3}{c}{\textbf{F1}} & \textbf{$\gamma$} & \textbf{$S_{\emptyset}$} \\
 &  & Hard & Soft & Hard & Soft & Hard & Soft & $\Delta$ &  &  \\
\midrule
Llama 3.3 & 0.336 & 0.070 & 0.243 & 0.063 & 0.219 & 0.066 & 0.230 & 0.164 & 0.092 & 0.343 \\
GPT-4o & 0.344 & 0.095 & 0.293 & 0.038 & 0.115 & 0.054 & 0.166 & 0.112 & 0.066 & 0.234 \\
Claude 3.7 & 0.460 & 0.110 & 0.274 & 0.096 & 0.239 & 0.103 & 0.255 & 0.152 & 0.155 & 0.113 \\
DeepS. R1 & 0.354 & 0.083 & 0.246 & 0.062 & 0.182 & 0.071 & 0.209 & 0.138 & 0.091 & 0.476 \\
o3-mini & 0.418 & \textbf{0.152} & \textbf{0.411} & 0.066 & 0.179 & 0.092 & 0.249 & 0.157 & 0.154 & \textbf{0.517} \\
Gem. 2-FT & \textbf{0.560} & 0.106 & 0.268 & \textbf{0.190} & \textbf{0.477} & \textbf{0.136} & \textbf{0.343} & 0.207 & \textbf{0.202} & 0.493 \\
\bottomrule
\end{tabular}
\caption{Evaluation of human and LLM annotations using \promptbase{} on the \proptask{}.}
\label{tab:results-propaganda-zeroshot-iaa}
\end{table*}



\begin{table}[htbp]
    \centering
    \small
\begin{tabular}{@{}lcccc@{}}
\toprule
\textbf{Annotator} & \textbf{Ann} & \textbf{Ann/Ex} & \textbf{w/o\%} & \textbf{Char/Ann} \\
\midrule
Human & 2981 & 2.5 & 28.8 & 50.3 \\
Llama 3.3 & 3214 & 2.7 & 7.4 & 65.5 \\
GPT-4o & 2284 & 1.9 & 4.8 & 66.3 \\
Claude 3.7 & 2865 & 2.4 & 22.5 & 57.2 \\
DeepS. R1 & 1387 & 1.2 & 44.2 & 56.8 \\
o3-mini & 1836 & 1.5 & 35.6 & 58.0 \\
Gem. 2-FT & 2517 & 2.1 & 28.9 & 54.3 \\
\bottomrule
\end{tabular}
\caption{Statistics of models and human annotators using \promptbase{} on \dttask{}. Ann=\# of annotations, Ann/Ex=ann. per example. w/o=\% ex. without annotations, Char/Ann=\# chars per ann.}
\label{tab:results-d2t-test-other}
\end{table}
\begin{table}[htbp]
    \centering
    \small
\begin{tabular}{@{}lcccc@{}}
\toprule
\textbf{Annotator} & \textbf{Ann.} & \textbf{Ann/Ex} & \textbf{w/o\%} & \textbf{Char/Ann} \\
\midrule
Human & 2090 & 0.7 & 66.0 & 14.5 \\
Llama 3.3 & 6361 & 2.3 & 6.2 & 17.4 \\
GPT-4o & 4866 & 1.7 & 7.0 & 15.9 \\
Claude 3.7 & 3782 & 1.4 & 30.6 & 15.9 \\
DeepS. R1 & 2586 & 0.9 & 36.3 & 15.1 \\
o3-mini & 3039 & 1.1 & 35.8 & 13.8 \\
Gem. 2-FT & 2181 & 0.8 & 50.0 & 15.2 \\
\bottomrule
\end{tabular}
\caption{Statistics of models and human annotators using \promptbase{} on \mttask{}. See \Cref{tab:results-d2t-test-other} for the legend.}
\label{tab:results-mt-zeroshot-sorted-all-other}
\end{table}

\small
\begin{table}[htbp]
    \centering
    \small
    \begin{tabular}{@{}llc@{\hspace{3pt}}c@{\hspace{3pt}}c@{\hspace{3pt}}c@{}}
        \toprule
        \textbf{Lang.} & \textbf{Annot.} & \textbf{Ann.} & \textbf{Ann/Ex} & \textbf{w/o\%} & \textbf{Char/Ann} \\
        \midrule
        \multirow{2}{*}{en-cs} & Model & 600 & 1.4 & 27.0 & 16.6 \\
                              & Human & 399 & 0.7 & 66.1 & 13.0 \\
        \midrule
        \multirow{2}{*}{en-es} & Model & 417 & 1.1 & 38.9 & 18.8 \\
                              & Human & 248 & 0.6 & 70.3 & 10.3 \\
        \midrule
        \multirow{2}{*}{en-hi} & Model & 396 & 1.3 & 26.2 & 19.0 \\
                              & Human & 222 & 0.5 & 71.2 & 10.7 \\
        \midrule
        \multirow{2}{*}{en-is} & Model & 563 & 1.9 & 14.3 & 15.7 \\
                              & Human & 752 & 2.5 & 18.3 & 16.6 \\
        \midrule
        \multirow{2}{*}{en-ja} & Model & 471 & 1.3 & 28.7 & 11.1 \\
                              & Human & 118 & 0.2 & 87.5 & 14.8 \\
        \midrule
        \multirow{2}{*}{en-ru} & Model & 500 & 1.3 & 25.9 & 18.2 \\
                              & Human & 287 & 0.7 & 58.7 & 19.4 \\
        \midrule
        \multirow{2}{*}{en-uk} & Model & 436 & 1.5 & 25.4 & 17.8 \\
                              & Human & 208 & 0.7 & 64.3 & 12.3 \\
        \midrule
        \multirow{2}{*}{en-zh} & Model & 420 & 1.2 & 34.6 & 7.2 \\
                              & Human & 171 & 0.2 & 85.1 & 6.6 \\
        \bottomrule
    \end{tabular}
    \caption{Statistics of models (averaged) and human annotators using \promptbase{} on the \mttask{} separately for each language. See \Cref{tab:results-d2t-test-other} for the legend.}
    \label{tab:results-mt-zeroshot-avg-other}
\end{table}

\begin{table}[htbp]
    \centering
    \small
\begin{tabular}{@{}lcccc@{}}
\toprule
\textbf{Annotator} & \textbf{Ann.} & \textbf{Ann/Ex} & \textbf{w/o\%} & \textbf{Char/Ann} \\
\midrule
Human & 1439 & 14.2 & 4.0 & 40.2 \\
Llama 3.3 & 574 & 5.7 & 3.0 & 92.0 \\
GPT-4o & 246 & 2.4 & 8.9 & 91.1 \\
Claude 3.7 & 803 & 8.0 & 7.9 & 58.5 \\
DeepS. R1 & 459 & 4.5 & 9.9 & 89.3 \\
o3-mini & 376 & 3.7 & 10.9 & 65.3 \\
Gem. 2-FT & 1864 & 18.5 & 3.0 & 54.1 \\
\bottomrule
\end{tabular}
\caption{Statistics of models and human annotators using \promptbase{} on the \proptask{}. See \Cref{tab:results-d2t-test-other} for the legend.}
\label{tab:results-propaganda-zeroshot-other}
\end{table}

\begin{table}[htbp]
\centering
\small
\begin{tabular}{@{}l|c@{\hspace{5pt}}c@{\hspace{5pt}}c@{\hspace{5pt}}c@{\hspace{5pt}}c|c@{\hspace{5pt}}c@{\hspace{5pt}}c@{\hspace{5pt}}c@{\hspace{5pt}}c@{}}
\toprule
& \multicolumn{5}{c|}{\textbf{Annotations}} & \multicolumn{5}{c}{\textbf{Explanations}} \\
Model & C & P & W & I & U & C & P & W & I & U \\
\midrule
Llama 3.3 & 7 & 1 & 2 & 8 & 0 & 6 & 0 & 1 & 10 & 1 \\
GPT-4o & 5 & 2 & 1 & 10 & 0 & 6 & 1 & 0 & 11 & 0 \\
Claude 3.7 & 7 & 2 & 3 & 6 & 0 & 9 & 2 & 0 & 7 & 0 \\
DeepSeek & 11 & 2 & 4 & 1 & 0 & 12 & 3 & 0 & 3 & 0 \\
o3-mini & 10 & 3 & 2 & 0 & 3 & 8 & 7 & 0 & 0 & 3 \\
Gemini 2 F-T & 12 & 4 & 1 & 1 & 0 & 12 & 5 & 1 & 0 & 0 \\
\midrule
Total & 52 & 14 & 13 & 26 & 3 & 53 & 18 & 2 & 31 & 4 \\
\bottomrule
\end{tabular}
\caption{Manual evaluation results for D2T-Eval domain. Categories for annotation and reason: C=Correct, P=Partially correct, W=Wrong category, I=Incorrect, U=Undecidable.}
\label{tab:manual-eval-d2t-eval}
\end{table}

\begin{table}[htbp]
\centering
\small
\begin{tabular}{@{}l|c@{\hspace{5pt}}c@{\hspace{5pt}}c@{\hspace{5pt}}c@{\hspace{5pt}}c|c@{\hspace{5pt}}c@{\hspace{5pt}}c@{\hspace{5pt}}c@{\hspace{5pt}}c@{}}
\toprule
& \multicolumn{5}{c|}{\textbf{Annotations}} & \multicolumn{5}{c}{\textbf{Explanations}} \\
Model & C & P & W & I & U & C & P & W & I & U \\
\midrule
Llama 3.3 & 9 & 1 & 2 & 5 & 1 & 8 & 2 & 1 & 6 & 1 \\
GPT-4o & 9 & 0 & 3 & 6 & 0 & 8 & 2 & 1 & 7 & 0 \\
Claude 3.7 & 9 & 2 & 3 & 4 & 0 & 9 & 3 & 3 & 3 & 0 \\
DeepSeek & 6 & 0 & 4 & 8 & 0 & 7 & 0 & 3 & 8 & 0 \\
o3-mini & 15 & 0 & 0 & 2 & 1 & 15 & 0 & 0 & 2 & 1 \\
Gemini 2 F-T & 7 & 3 & 2 & 6 & 0 & 9 & 2 & 1 & 6 & 0 \\
\midrule
Total & 55 & 6 & 14 & 31 & 2 & 56 & 9 & 9 & 32 & 2 \\
\bottomrule
\end{tabular}
\caption{Manual evaluation results for Propaganda domain. See \Cref{tab:manual-eval-d2t-eval} for the legend.}
\label{tab:manual-eval-propaganda}
\end{table}

\end{document}